\documentclass[twocolumn]{article}
\usepackage{arxiv}
\usepackage{graphicx}
\usepackage{color}
\usepackage{xcolor}
\usepackage{latexsym}
\usepackage{amsmath,amssymb}
\usepackage{multicol, multirow}
\usepackage{url}
\usepackage{here}

\def\Underline{\setbox0\hbox\bgroup\let\\\endUnderline}
\def\endUnderline{\vphantom{y}\egroup\smash{\underline{\box0}}\\}
\def\|{\verb|}
\def\figref#1{Figure \ref{#1}}
\def\tabref#1{Table \ref{#1}}
\def\secref#1{Section \ref{#1}}
\def\myeqref#1{Equation \eqref{#1}}

\def\l#1{\underline{#1}}
\def\g#1{}
\def\ours{FlowFM}

\title{High-Performance Self-Supervised Learning \\by Joint Training of Flow Matching}
\author{
    Kosuke Ukita \\
        Kyushu Institute of Technology \\
        \texttt{ukita.kosuke299@mail.kyutech.jp} \\
    \And
    Tsuyoshi Okita \\
        Kyushu Institute of Technology \\
        \texttt{tsuyoshi@ai.kyutech.ac.jp} \\
}

\begin{document}
\maketitle

\begin{abstract}
Diffusion models can learn rich representations during data generation, showing potential for Self-Supervised Learning (SSL), but they face a trade-off between generative quality and discriminative performance. Their iterative sampling also incurs substantial computational and energy costs, hindering industrial and edge AI applications. To address these issues, we propose the Flow Matching-based Foundation Model (FlowFM), which jointly trains a representation encoder and a conditional flow matching generator. This decoupled design achieves both high-fidelity generation and effective recognition. By using flow matching to learn a simpler velocity field, FlowFM accelerates and stabilizes training, improving its efficiency for representation learning.
Experiments on wearable sensor data show FlowFM reduces training time by 50.4\% compared to a diffusion-based approach. On downstream tasks, FlowFM surpassed the state-of-the-art SSL method (SSL-Wearables) on all five datasets while achieving up to a 51.0x inference speedup and maintaining high generative quality.
The implementation code is available at 
\url{https://github.com/Okita-Laboratory/jointOptimizationFlowMatching}.
\end{abstract}

\section{Introduction} \label{section:introduction}
Diffusion models have recently emerged as a leading class of generative models, initially achieving remarkable quality in image generation and now driving breakthroughs across diverse applications like video generation~\cite{zhang20244diffusion}, time-series forecasting~\cite{meijer2024rise}, speech synthesis~\cite{shen2023naturalspeech}, and scientific domains such as drug discovery~\cite{huang2024dual} and robotics~\cite{shaoul2024multi}.

A pivotal discovery fueling their success is that in generating high-quality data, these models incidentally learn powerful discriminative features, suggesting their potential as a next-generation framework for self-supervised learning (SSL)~\cite{chen2024deconstructing,xiang2023denoising}.

However, applying diffusion models to representation learning presents a fundamental dilemma. Repurposing existing generative models (e.g., DDAE~\cite{xiang2023denoising}) yields representations not optimized for recognition tasks~\cite{yang2023diffusion}, while redesigning them for representation learning (e.g., l-DAE~\cite{chen2024deconstructing}, SODA~\cite{hudson2024soda}) often sacrifices their powerful generative capabilities. This stems from an inherent trade-off between generative and discriminative quality~\cite{chen2024deconstructing,dombrowski2024trade}, as the goal of faithfully reproducing fine details conflicts with learning representations that are invariant to task-irrelevant noise.

A promising solution is to decouple the architecture into a representation encoder and a generative network trained jointly. However, implementing this with diffusion models introduces another major challenge: the immense computational cost of their iterative training and inference processes. To overcome these dual challenges of the performance trade-off and high computational cost, we propose \ours{} (Flow Matching-based Foundation Model). \ours{} is built on flow matching, a recent paradigm that achieves diffusion-like generative quality with a simpler and faster synthesis process. Our core idea is to integrate the decoupled recognition-generation architecture within this computationally efficient framework. By uniting architectural separation with computational efficiency, \ours{} achieves high generative and discriminative performance without compromise, creating a more versatile and powerful foundation model.

The contributions of this paper are as follows:
\begin{itemize}
\item We propose \ours{}, a novel foundation model that directly utilizes the generative process of flow matching for representation learning.
\item We introduce an efficient method for jointly training a representation encoder and a velocity field prediction network using the flow matching objective function.
\item We demonstrate that our model significantly outperforms the state-of-the-art SSL baseline, SSL-Wearables, on human activity recognition tasks.
\item Through a Text-to-Signal generation task, we show that the learned representations capture not only discriminative information but also generative structure.
\end{itemize}
This paper is organized as follows. In \secref{section:related_work}, we review related work. In \secref{section:background}, we cover the necessary background, and in \secref{section:methods}, we describe our proposed method in detail. Next, in \secref{section:experiments}, we present the experimental results. Finally, in \secref{section:conclusion}, we conclude the paper.

\section{Related Work} \label{section:related_work}

\subsection{Self-Supervised Representation Learning}
SSL learns general-purpose feature representations by generating supervisory labels from the data itself, without the need for manual annotations. Modern SSL has been primarily driven by two major paradigms.

\paragraph{Masked Modeling}
Masked modeling involves training a model to predict a masked portion of an input sequence from its surrounding context. This approach revolutionized natural language processing with the introduction of BERT~\cite{devlin2019bert} and has since achieved tremendous success in diverse modalities, such as with Masked Autoencoders (MAE)~\cite{he2022masked} in computer vision. The essence of this method is to compel the model to acquire an understanding of global context and semantic representations by solving the task of restoring local information.

\paragraph{Contrastive Learning}
Contrastive learning defines different augmentations of the same data as "positive pairs" and augmentations from different data as "negative pairs." The model is trained to maximize the similarity between positive pairs and minimize it between negative pairs in the representation space. Representative methods include MoCo~\cite{he2019moco}, SimCLR~\cite{chen2020simple}, and DINO~\cite{caron2021emerging}. This paradigm is particularly known for learning features that exhibit excellent linear separability in downstream tasks.

\subsection{Representation Learning Using Diffusion Models}
The powerful data modeling capabilities of diffusion models have opened up a new research frontier in SSL. Approaches in this area can be broadly categorized as follows.

\paragraph{Reusing Pretrained Diffusion Models}
Pioneering work~\cite{xiang2023denoising} discovered that the intermediate representations of diffusion models pretrained for generative tasks unintentionally contain high-quality discriminative features. DDAE~\cite{xiang2023denoising} demonstrated a way to leverage existing models as zero-cost feature extractors. RepFusion~\cite{yang2023diffusion} further proposed a sophisticated method that combines knowledge distillation and reinforcement learning to dynamically extract optimal features tailored to a specific task. Diffusion Classifier~\cite{li2023diffusion} extended this to zero-shot classification tasks. However, since these approaches rely on existing models optimized for generation, their representations are not always optimal for downstream tasks.

\paragraph{Specialization and Simplification for Representation Learning}
An alternative approach is to rethink the model architecture itself for the purpose of representation learning. l-DAE~\cite{chen2024deconstructing} deconstructed and simplified diffusion models into their essential component, the Denoising Autoencoder, showing that the core of representation learning lies in denoising within the latent space. SODA~\cite{hudson2024soda} aimed to engineer semantically disentangled representations by introducing an information bottleneck into the architecture and imposing specific self-supervised tasks. While these studies offer important insights, they tend to sacrifice the powerful generative capabilities of diffusion models in pursuit of discriminative performance.

\subsection{Flow Matching for Generative Modeling}
To address the computational cost issues of diffusion models, particularly their slow sampling speed, flow matching~\cite{lipman2022flow, liu2022flow} has recently been proposed as a new paradigm for generative modeling. Flow matching directly learns the velocity field of an Ordinary Differential Equation (ODE) that defines a continuous flow from a simple probability distribution to the data distribution. This approach aims for efficient, high-quality data generation, potentially in a single step, without the need for iterative sampling like in diffusion models. Conditional Flow Matching~\cite{tong2023improving, tong2023simulation} provided a computationally tractable objective function for training neural networks by targeting simple velocity fields conditioned on individual sample pairs, which simplifies the otherwise difficult problem of learning the average velocity field.
However, research on flow matching to date has predominantly focused on high-quality data generation itself. The discriminative capabilities of the representations learned in the process have not been a central focus, and attempts to apply them to build foundation models are, to the best of our knowledge, largely non-existent. This paper ventures into this unexplored territory.

\section{Background} \label{section:background}
In \secref{section:background}, we provide an overview of two fundamental technologies that underpin our proposed method: Latent Diffusion Models and Flow Matching.

\subsection{Latent Diffusion Models}
In contrast to earlier diffusion models such as DDPM, score-based models, and SDE-based models~\cite{ho2020denoising,song2020score}, Latent Diffusion Models (LDMs)~\cite{rombach2022high} construct the diffusion process in a compressed latent space, using methods like VAEs to encode pixel-space data. This approach significantly improves both the quality and speed of generation. Furthermore, the Diffusion Transformer (DiT)~\cite{peebles2023scalable} replaced the commonly used U-Net backbone with a Transformer module, demonstrating enhancements in generation quality and scalability. In this work, we use DiT as our base architecture.

\subsection{Flow Matching}
Conditional Flow Matching (CFM) resolves the computationally challenging problem of learning a distribution's average velocity field by targeting simpler fields conditioned on individual sample pairs. This approach yields a computationally tractable objective function for training neural networks.
In this framework, we first draw a source sample $x_0 \in \mathbb{R}^d$ from a prior distribution $p_0$ that is easy to sample from, and a target sample $x_1 \in \mathbb{R}^d$ from the training data distribution $p_1$. Next, using a time variable $t \in [0, 1]$, a point $x_t$ on the probability path connecting these two points is constructed, most commonly via linear interpolation as shown in \myeqref{formula:xt}.
\begin{equation}
\label{formula:xt}
x_t = (1-t) x_0 + t x_1
\end{equation}
The velocity field model $v_\theta$ is trained to predict the target velocity, denoted as $u_t(x_t|x_1)$, for any point $x_t$ on this path. The CFM loss minimizes the squared error between the model's prediction and this target velocity, as expressed in \myeqref{formula:cfmloss}~\cite{lipman2024flow}.
\begin{equation}
\label{formula:cfmloss}
\mathcal{L}_{\textit{CFM}}(\theta)
= \mathbb{E}_{t, x_0, x_1} \left[ \parallel v_\theta(x_t,t) - u_t(x_t|x_1) \parallel^2 \right]
\end{equation}

\section{Proposed Method} \label{section:methods}
\begin{figure*}[h]
    \begin{center}
    \includegraphics[width=0.75\textwidth]{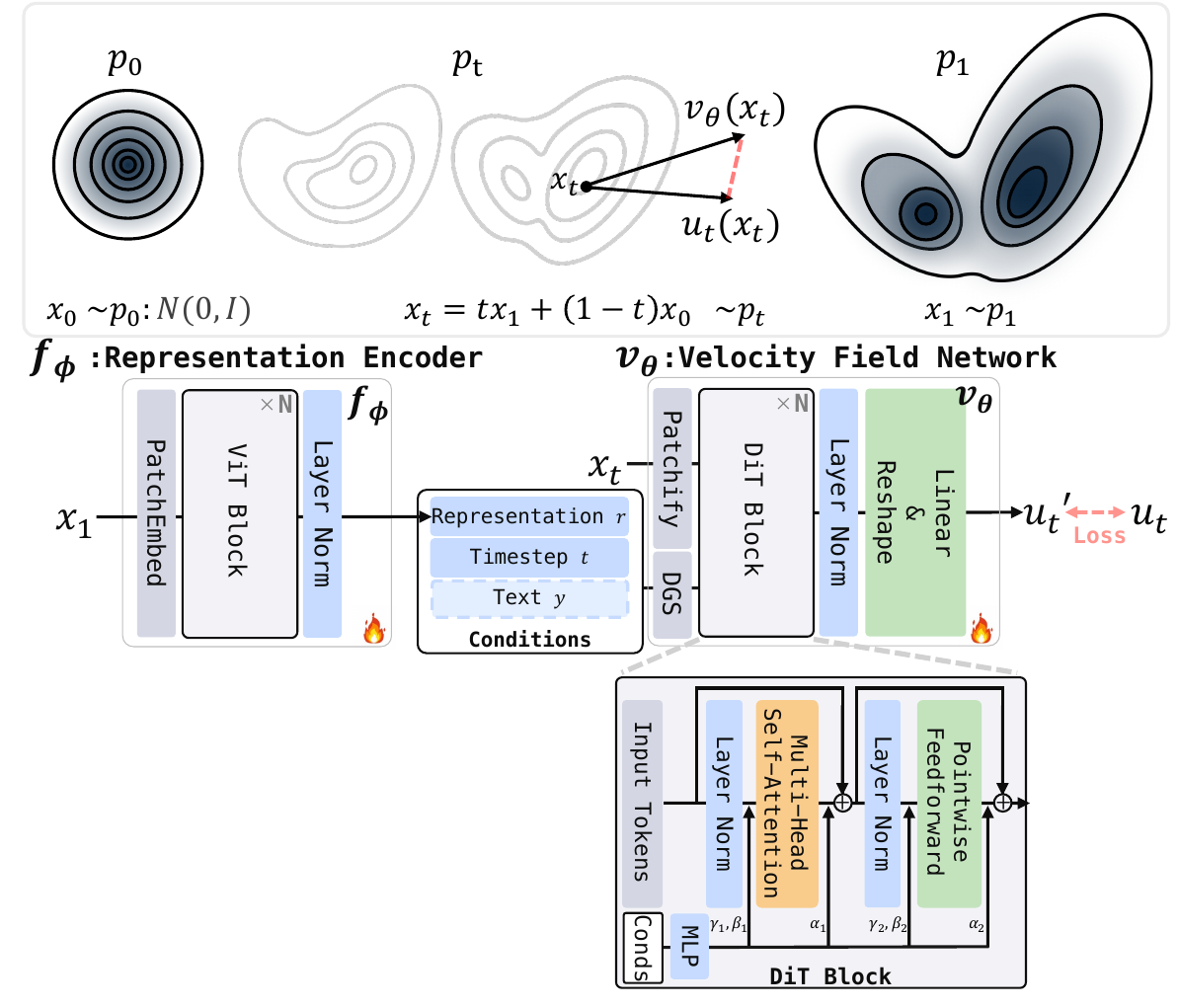}
    \caption{\textbf{Overall architecture of \ours{} during pre-training.} The model consists of (left) Representation Encoder $f_\phi$ and (right) Velocity Field Network $v_\theta$. $f_\phi$ extracts representation $r$ from input $x_1$ and supplies it as a condition to $v_\theta$. $v_\theta$ takes point $x_t$ on the probability path and conditions ($r$, time $t$, text $y$) and is jointly trained to predict the target velocity field $u_t$. The DGS module improves robustness by preventing over-reliance on the representation by randomly masking $r$ during training.}
    \label{fig:flowfm}
    \end{center}
\end{figure*}

In \secref{section:methods}, we detail \ours{}, a novel foundation model that utilizes flow matching. Our method follows the conventional two-stage process of self-supervised learning~\cite{devlin2019bert,caron2021emerging}, which involves pre-training on an unlabeled dataset and subsequent adaptation to downstream tasks. We build this framework using flow matching, but diverge from typical conditional generation approaches used in tasks like Text-to-Image, which often rely on large-scale annotated datasets. In contrast, to mitigate annotation costs, we pre-train our model by learning a representation-conditioned generation process that does not require any labels. The representations obtained through this process are then leveraged for downstream applications.

\subsection{Pre-training}
The training architecture of the proposed \ours{} during pre-training is shown in \figref{fig:flowfm}. \ours{} is primarily composed of two components: a "Representation Encoder $f_{\phi}(\cdot)$" and a "Velocity Field Network $v_{\theta}(\cdot)$". In this paper, we specifically refer to the neural network that takes input data $x_1$ and outputs a representation vector $r \in \mathbb{R}^d$ as the Representation Encoder, as shown in \myeqref{formula:encode}.
\begin{equation}
    \label{formula:encode}
    r = f_\phi(x_1)
\end{equation}
The representation obtained from the Representation Encoder is fed as one of the conditions to the Velocity Field Network.
Specifically, our work deals with conditional generation, producing target data $x_1$ from a condition $c$ such as text. To this end, we first extend the CFM loss in \myeqref{formula:cfmloss} to a framework for conditional generation. With the goal of learning the flow from a prior distribution $p_0$ to a target distribution $p_1(x_1|c)$ given a condition $c$, the CFM loss for conditional generation is given in \myeqref{formula:cfmcondloss}.
\begin{align}
\label{formula:cfmcondloss}
\mathcal{L}&_{\textit{CFM, cond}}(\theta) \notag \\
&= \mathbb{E}_{t, x_0, (x_1, c)} \left[ \parallel v_\theta(x_t,t,c) - u_t(x_t|x_1, c) \parallel^2 \right]
\end{align}
The loss for \ours{}, which learns conditional generation conditioned on the active representation $r$ obtained from the encoder, is shown in \myeqref{formula:flowfmloss}.
\begin{align}
&\mathcal{L}_{\textit{FlowFM}}(\theta,\phi) \notag\\
&= \mathbb{E}_{t, x_0, x_1} \left[ \parallel v_\theta(x_t,t,r) - u_t(x_t|x_1, r) \parallel^2 \right] \notag\\
&= \mathbb{E}_{t, x_0, x_1} \left[ \parallel v_\theta(x_t,t,f_\phi(x_1)) - u_t(x_t|x_1, f_\phi(x_1)) \parallel^2 \right] 
\label{formula:flowfmloss}
\end{align}
The velocity field network $v_\theta$ is conditioned on the representation $r$, which depends on parameters $\phi$, and is trained by optimizing with respect to parameters $\theta$ and $\phi$.
Although several methods have been proposed for training a representation-conditioned generation process in diffusion models~\cite{rombach2022high,bordes2022high}, they all use representations from a pretrained, fixed encoder, making the conditioning information static and not a target of learning. In contrast, our method is different in that it trains the Representation Encoder concurrently with the generation process. The representation in the condition changes dynamically and is updated. By training the representation via gradients from the flow matching generation process, it is expected to acquire not only generative capabilities but also advanced recognition abilities.
\begin{align}
\mathcal{L}_{\textit{FlowFM}}&(\theta,\phi) \notag\\
&= \mathbb{E}_{t, x_0, x_1} \left[ \parallel v_\theta^\phi(x_t,t) - u_t^\phi(x_t|x_1) \parallel^2 \right]
\label{formula:flowfmfieldloss}
\end{align}
The \ours{} loss shown in \myeqref{formula:flowfmloss} can be rewritten as \myeqref{formula:flowfmfieldloss}, which offers the following interpretation: the parameters $\phi$ of the representation encoder $f_\phi$ do not merely provide a static condition, but dynamically parameterize both the model's predicted velocity field $v_\theta$ and the target velocity field $u_t$. This means that our framework does not solve a fixed, complex generation problem. Instead, $\phi$ actively designs a "path $u_t^\phi$" that simplifies the problem itself, and the framework simultaneously performs this problem simplification and derives its solution.

\paragraph{Velocity Field Network}
The velocity field network in this study is based on the DiT~\cite{peebles2023scalable} architecture, which has demonstrated high performance in recent image generation tasks. However, since DiT is originally designed for 2D image data, we have modified the input part to process 1D sensor data. Specifically, within the patch embedding layer at the input of DiT, we replaced it with a 1D convolutional layer. This converts the continuous time-series signal into a sequence of tokens that can be processed by the Transformer blocks. The configuration of the Transformer blocks follows the original DiT.

\paragraph{Representation Encoder}
Similarly, the representation encoder adopts a Transformer-based ViT~\cite{dosovitskiy2020image} architecture. We modified its patch embedding layer with a 1D convolutional layer to enable processing of 1D sensor data. It extracts a single feature vector from the input 1D sensor data, which serves as the representation vector.

\subsubsection{Conditioning Mechanisms}
In our method, we provide multiple pieces of information as conditions to the velocity field network to control and stabilize the generative process for conditional generation.
During pre-training, the conditioning information consists of two types: (1) the representation vector $r$ output from the representation encoder, which is the core of our proposed method, and (2) the timestep $t$. The timestep $t$ is vectorized using a sinusoidal position encoding. These two condition vectors are combined element-wise and then integrated into each DiT block via \textit{adaLN-Zero}\footnote{The conditioning mechanism employed in DiT.}. Furthermore, considering various downstream tasks, this module is designed to allow for the addition of multiple arbitrary conditions, such as text prompts.

\paragraph{Dynamic Guidance Switching}
To enhance the model's versatility and improve performance on downstream tasks, we employ a strategy of intentionally masking the condition during training, which we name Dynamic Guidance Switching (DGS). Specifically, at each training step, the representation vector $r$ is replaced with a zero vector with a certain probability (50\% in this study). DGS, which is not used in the original DiT~\cite{peebles2023scalable} architecture and is our own design, enables a single model to learn both unconditional and conditional generation. It also acts as an implicit regularization on the representations acquired by the encoder.
When the guiding information provided by the representation is masked, the generation process must maintain high generative capability autonomously without relying on that information. This prevents the generation process from becoming overly dependent on the representation and deteriorating in quality. Then, when the representation is provided as a condition, it must offer essential information that facilitates learning to contribute to the training of the generation process. Therefore, masking the conditioning representation is expected to be effective. To verify this, we conducted a comparative experiment between training without DGS and with DGS applied, where the representation was randomly masked with a 50\% probability.
As shown in \tabref{table:result_repcondclassifier}, the introduction of DGS consistently improved classification performance on the HAR task compared to not using it. This result demonstrates that masking the representation acts as a form of regularization, enabling the representation encoder to learn more robust and essential features.

\begin{table}[h]
    \centering
    \caption{\textbf{Effect of masking strategy}: Comparison of fine-tuning accuracy with 0\% vs 50\% mask probabilities. The 50\% setting (DGS) consistently yields higher accuracy across all datasets, demonstrating that random masking helps learn more robust representations.}
    \label{table:result_repcondclassifier}
    \scalebox{0.8}{
    \begin{tabular}{c|cccc} \hline\hline
        Mask prob. & ADL & Opportunity & PAMAP2 & REALWORLD \\ \hline
        0\%   & .9370 & .7775 & .8868 & .9004 \\
        50\%  & \l{.9449} & \l{.7974} & \l{.8920} & \l{.9112} \\
        \hline
    \end{tabular}
    }
\end{table}

\subsection{Downstream Tasks}
\subsubsection{Human Activity Recognition Task}
To quantitatively evaluate the quality of the representations acquired through pre-training, we perform a Human Activity Recognition (HAR) task. This task utilizes the representation encoder trained during pre-training. We evaluate using two methods: transfer learning, where the pretrained encoder is frozen and only a linear classifier is trained, and fine-tuning, where all parameters are updated using the pretrained encoder as initialization.

\subsubsection{Text-to-Signal Task}
To verify that the representations acquired by \ours{} pre-training capture not just discriminative information but also richer, semantic structures, we conduct a signal generation task conditioned on text descriptions.
Signal generation in this task is based on the velocity field network trained during pre-training. While the generative model obtained from \ours{} pre-training is capable of zero-shot unconditional and representation-conditioned generation, it does not inherently possess the ability to generate from text descriptions due to the self-supervised learning setup. Therefore, we tune the network to generate high-quality time-series signals conditioned on text descriptions, using the learned network weights as an initial state. 
\begin{figure}[t]
    \begin{center}
    \includegraphics[width=0.48\textwidth]{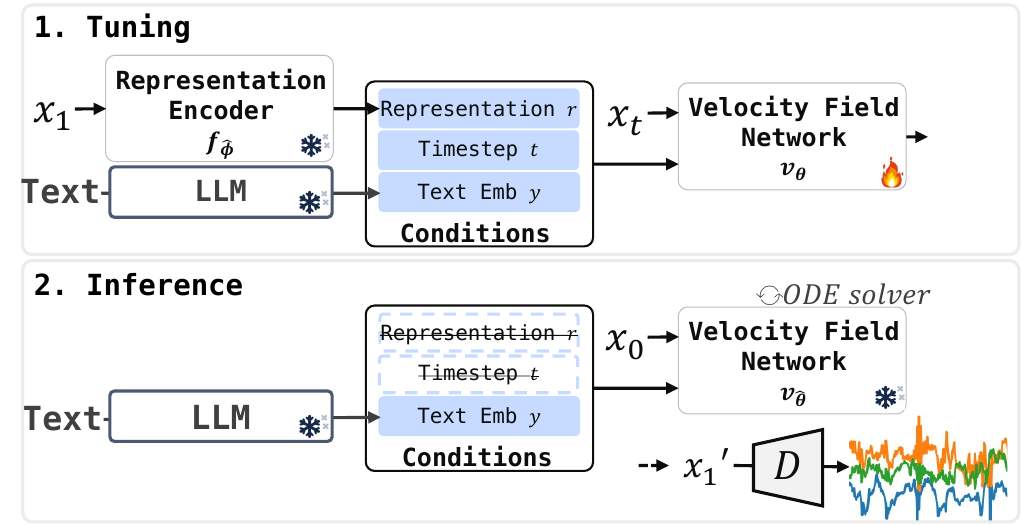}
    \caption{\textbf{Text-to-Signal workflow}: (1) In tuning, the network $v_{\theta}$ is fine-tuned conditioned on both representation $r$ and text embedding $y$ with frozen components. (2) In inference, the model generates signals $x'_1$ from noise $x_0$ via an ODE solver, conditioned solely on text $y$.}
    \label{fig:tts_model}
    \end{center}
\end{figure}
\figref{fig:tts_model} illustrates the workflow for tuning and signal generation during inference for the Text-to-Signal task.
The tuning process is achieved by providing both the input text description and the representation as conditions. In diffusion models, it has been reported that incorporating representations into the conditional generation process improves generation quality~\cite{rombach2022high,bordes2022high}. Therefore, we adopt an approach where we combine text embeddings with representations obtained from a fixed representation encoder and provide them as a condition. As shown in \tabref{tab:fid}, this composite conditioning method consistently yields lower FID scores compared to using only text embeddings, demonstrating the generation of higher-fidelity signals. This result suggests that the representations acquired during pre-training richly capture the detailed statistical properties of the data and function as useful prior knowledge that further enhances the performance of the generative model when combined with high-level semantic information like text.

During inference, signals are generated solely from text descriptions. The text description input by the user is converted into a text embedding via an LLM, which serves as part of the condition for the denoising network. Unlike the tuning phase, this is designed not to require a representation.

\begin{table}[h]
\begin{center}
    \caption{\textbf{Impact of conditioning methods}: Comparison of generation quality (FID, Precision, Recall) on PAMAP2 and REALWORLD. Combining text with learned representations (Text + Rep) consistently outperforms Text-Only conditioning, demonstrating that incorporating representations significantly enhances signal fidelity.}
    \label{tab:fid}
    \scalebox{0.82}{
    \begin{tabular}{cc|ccc} \hline\hline
        Dataset & Conditioning & FID$\downarrow$ & Precision$\uparrow$ & Recall$\uparrow$ \\ \hline
        \multirow{2}{*}{PAMAP2} 
            & Text-Only  & 20.24 & 0.4843 & 0.5188 \\
            & Text + Rep & \l{9.991}& \l{0.6300} & \l{0.6597} \\ \hline
        \multirow{2}{*}{REALWORLD} 
            & Text-Only  & 12.33 & 0.2486 & 0.2480 \\
            & Text + Rep & \l{8.261} & \l{0.6838} & \l{0.6925} \\ \hline
    \end{tabular}
    }
\end{center}
\end{table}

\begin{table*}[h]
    \centering
    \caption{\textbf{HAR classification performance}: Comparison of FlowFM against state-of-the-art baselines (e.g., SSL-Wearables) across five datasets. FlowFM consistently achieves the highest accuracy and F1-scores in both transfer learning and fine-tuning settings.}
    \label{table:classification}
    \scalebox{0.8}{
    \begin{tabular}{c|cc|cc|cc|cc|cc} \hline\hline
        \multirow{2}{*}{Method}               & 
              \multicolumn{2}{c}{ADL}       &
              \multicolumn{2}{c}{Opportunity} &
              \multicolumn{2}{c}{PAMAP2}      &
              \multicolumn{2}{c}{REALWORLD}   &
              \multicolumn{2}{c}{WISDM}       \\
              & acc. & f1 & acc. & f1 & acc. & f1 & acc. & f1 & acc. & f1 \\ \hline

        \multicolumn{11}{c}{Transfer learning} \\ \hline

        SSL-Wearables \cite{yuan2024self} &
        - & .7540 & - & .5470 & - & .7250 & - & .7710 & - & .7230 \\

        1D-DINO &
        .8835 & .8691 & .7557 & .7682 & .8111 & .7981 & .8124 & .8243 & .7747 & .7716 \\
        
        SENvT-u4~\cite{okita2023} &
        .8394 & .7936 & .7103 & .6826 & .7157 & .6912 & .7470 & .7512 & .6667 & .6618 \\
        
        SENvT-u7~\cite{okita2023} &
        .8472 & .7849 & .7150 & .6909 & .7098 & .6869 & .7628 & .7692 & .6819 & .6766 \\
        
        \hline
        
        \textbf{DiffFM (ours)} &
        .9055 & .8813 & .7787 & .7760 & .8171 & .8090 & .8614 & .8741 & .8261 & .8224 \\

        \textbf{\ours{} (ours)} &
        \l{.9370} & \l{.9168} & \l{.7881} & \l{.7792} & \l{.8467} & \l{.8373} & \l{.8683} & \l{.8832} & \l{.8337} & \l{.8322} \\
        \hline\hline
        
        \multicolumn{11}{c}{Fine-tuning} \\ \hline

        \g{Random Forest\footnote[1]{}} &
        \g{.8346} & \g{.6830} & \g{.6814} & \g{.4260} & \g{.6602} & \g{.5467} & \g{.6618} & \g{.5712} & \g{.4630} & \g{.4224} \\
        
        \g{Transformer scratch\footnote[2]{}} &
        \g{.8503} & \g{.8230} & \g{.6276} & \g{.6209} & \g{.7682} & \g{.7594} & \g{.8554} & \g{.8663} & \g{.8219} & \g{.8194} \\

        BERT \cite{devlin2019bert} &
        .8583 & .8553 & .7272 & .7245 & .7561 & .7527 & .7807 & .7819 & .7923 & .7924 \\

        SSL-Wearables \cite{yuan2024self} &
        - & .8290 & - & .5950 & - & .7890 & - & .7920 & - & .8100 \\

        1D-DINO &
        .9024 & .8780 & .7656 & .7741 & .8341 & .8223 & .8878 & .8978 & .8693 & .8684 \\
        
        SENvT-u4~\cite{okita2023} &
        .9087 & .8780 & .7489 & .7411 & .8488 & .8454 & .9015 & .9124 & .8693 & .8679 \\
        
        SENvT-u7~\cite{okita2023} &
        .8992 & .8684 & .7429 & .7448 & .8477 & .8445 & .9058 & .9163 & .8875 & .8867 \\

        1D-Diffusion Classifier &
        .8818 & .8515 & .7693 & .7526 & .8763 & .8768 & .8959 & .8574 & .8076 & .8075 \\ \hline
        
        \textbf{DiffFM (ours)} &
        .9291 & .9118 & .7857 & .7782 & .8902 & \l{.8877} & .9084 & .9197 & .8839 & .8832 \\

        \textbf{\ours{} (ours)} &
        \l{.9449} & \l{.9286} & \l{.7974} & \l{.7925} & \l{.8920} & \l{.8877} & \l{.9112} & \l{.9211} & \l{.8918} & \l{.8910} \\
        \hline
    \end{tabular}
    }
    \begin{itemize}
        \setlength{\leftskip}{5pt}
        \footnotesize{
        \item[*1] Random Forest: 900-d input (concatenated x, y, z), depth 5, 10 trees.
        \item[*2] Transformer: depth 6, embedding dim 128, 4 heads, learning rate 0.001, 50 epochs.
        }
    \end{itemize}
\end{table*}

\begin{figure}[h]
    \begin{center}
    \includegraphics[width=0.45\textwidth]{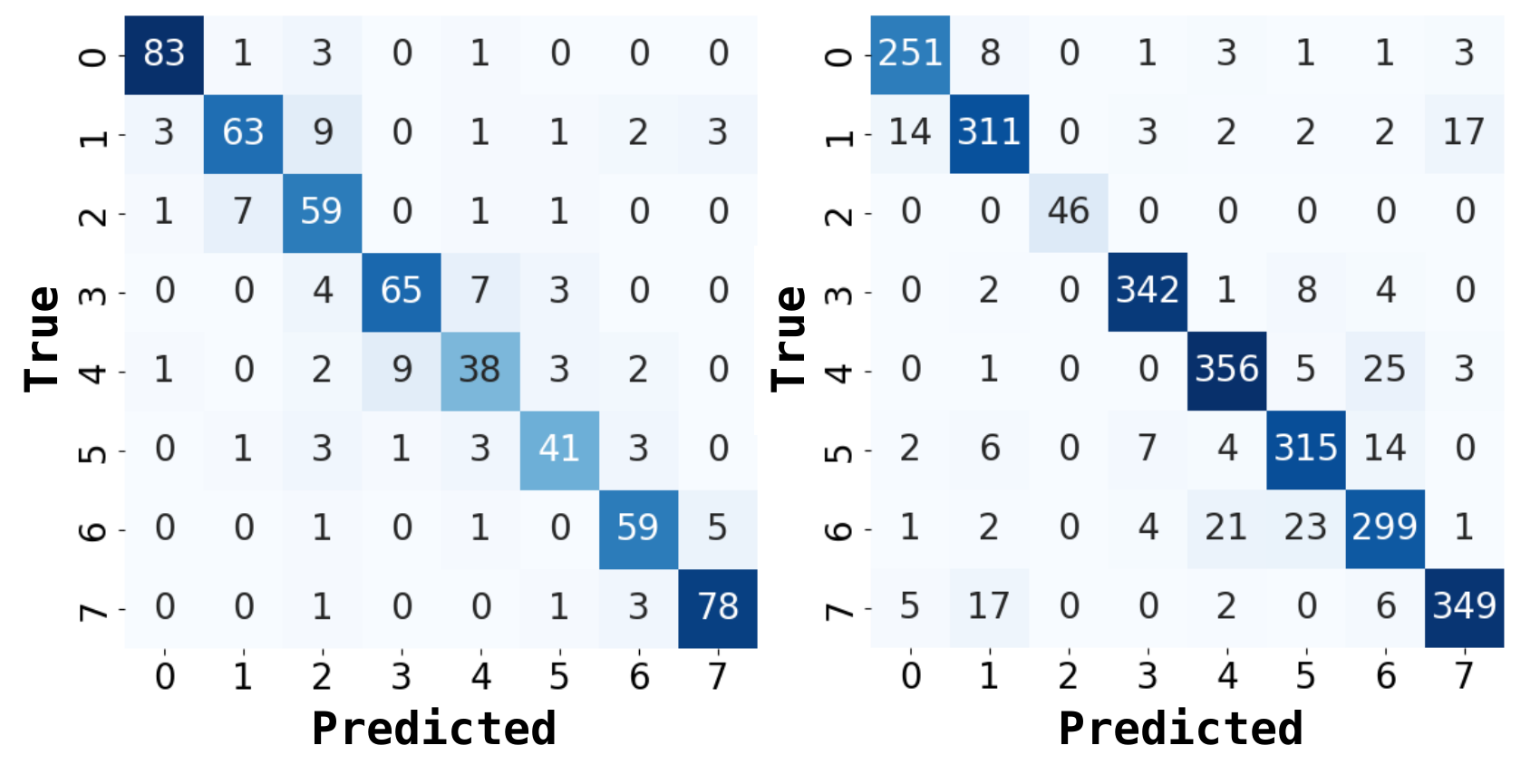}
    \caption{\textbf{Confusion matrices on HAR tasks}: The left panel displays transfer learning results on the PAMAP2 dataset, while the right panel shows fine-tuning results on the REALWORLD dataset. The strong diagonal patterns demonstrate FlowFM's high classification accuracy across diverse activity classes.}
    \label{fig:pamap_cm}
    \end{center}
\end{figure}

\section{Experiments} \label{section:experiments}
\begin{figure}[h]
    \begin{center}
    \includegraphics[width=0.5\textwidth]{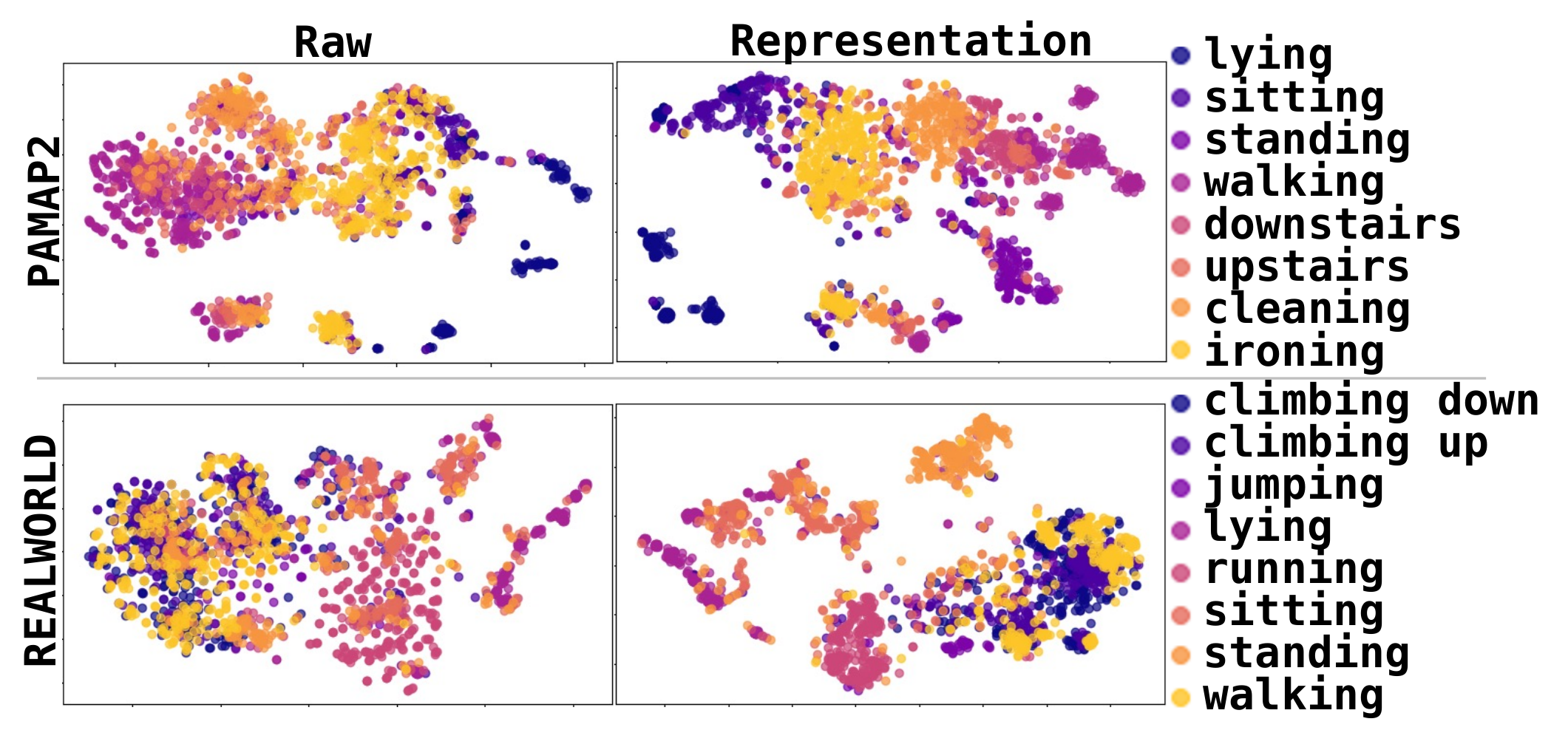}
    \caption{\textbf{t-SNE visualization of learned representations}: Comparison between raw data (left) and FlowFM representations (right) on PAMAP2 and REALWORLD datasets. The learned representations successfully disentangle activity clusters that overlap in the raw data space, demonstrating high-quality feature separation without supervision.}
    \label{fig:rep_map}
    \end{center}
\end{figure}
In \secref{section:experiments}, we describe the datasets, experimental setup, and results. First, in \secref{section:datasets}, we explain the datasets used in our experiments. Next, in \secref{section:eval_har}, we provide a quantitative evaluation of the quality of the learned representations through a human activity recognition task. In \secref{section:eval_gen_sensor}, we present a qualitative evaluation through a Text-to-Signal generation task conditioned on text descriptions. Finally, in \secref{section:efficient}, we report the results regarding the computational cost of \ours{}.
To clearly demonstrate the effectiveness of our proposed \ours{}, we designed a diffusion model-based counterpart for direct comparison. We name this model the Diffusion-based Foundation Model (DiffFM). It employs the same joint training architecture of a representation encoder and a generative network as FlowFM, but with the generation mechanism replaced by a standard diffusion model. By directly comparing the performance and computational costs of FlowFM and DiffFM, we quantitatively evaluate the improvements in recognition performance and efficiency achieved by FlowFM.
\begin{figure*}[h]
    \begin{center}
    \includegraphics[width=0.98\textwidth]{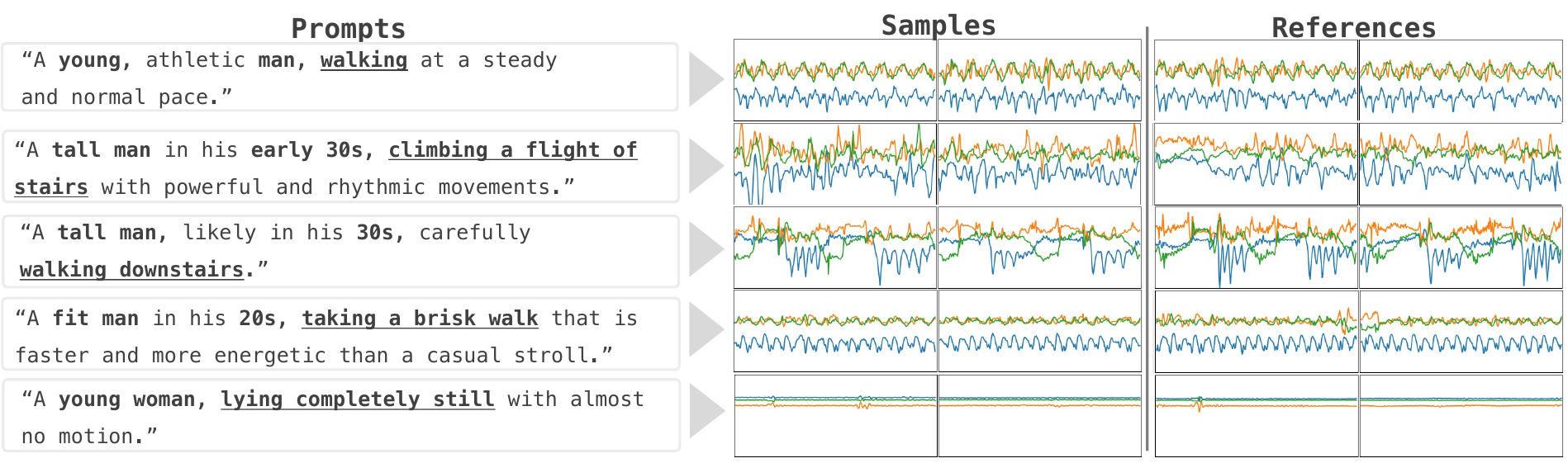}
    \caption{\textbf{Text-to-Signal generation examples}: Generated 3-axis accelerometer signals (Samples) conditioned on text prompts are shown alongside reference data (References). The model accurately translates semantic descriptions—such as activity type, intensity, and subject characteristics—into realistic time-series patterns, capturing distinct motion signatures like "walking" versus "climbing stairs."}
    \label{fig:text_to_sensor}
    \end{center}
\end{figure*}
\subsection{Datasets} \label{section:datasets}

\begin{table}[h]
    \begin{center}
    \caption{\textbf{Datasets}}
    \label{table:dataset}
    \scalebox{0.9}{
        \begin{tabular}{cccc} \hline\hline
          Dataset & Subjects & Samples & Classes \\\hline
          Capture-24 \cite{willetts2018statistical} & 152 & 1.3M &  4 \\ \hline
          ADL \cite{bruno2013analysis}            &   7 & 0.6K &  5 \\
          Opportunity \cite{roggen2010collecting}   &   4 & 3.9K &  4 \\
          PAMAP2 \cite{reiss2012introducing}        &   8 & 2.9K &  8 \\
          REALWORLD \cite{sztyler2016body}          &  14 &  12K &  8 \\
          WISDM \cite{weiss2019smartphone}          &  46 &  28K & 18 \\ \hline
        \end{tabular}
    }
    \end{center}
\end{table}

In this experiment, we use 3-axis accelerometer data from wrist-worn IMU sensors. Following Yuan et al.~\cite{yuan2024self}, we segment the signals into fixed-length sliding windows and treat them as independent inputs. All data used in the experiments are resampled to a frequency of 30 Hz using a 10-second sliding window. Details of the datasets are shown in \tabref{table:dataset}. The pre-training data consists of the Capture-24 dataset~\cite{willetts2018statistical}, which we use as an unlabeled dataset. This dataset contains a total of 1,387,255 sliding windows. For the downstream tasks, we use five datasets: ADL~\cite{bruno2013analysis}, Opportunity~\cite{roggen2010collecting}, PAMAP2~\cite{reiss2012introducing}, REALWORLD~\cite{sztyler2016body}, and WISDM~\cite{weiss2019smartphone}.

\subsection{Human Activity Recognition Results} \label{section:eval_har}
We conducted an HAR task to quantitatively evaluate the quality of the representations learned during pre-training. The Accuracy (acc.) and F1-score (f1) results are shown in \tabref{table:classification}.
We compared our method against several baselines: 
SSL-Wearables~\cite{yuan2024self}, a state-of-the-art multi-task SSL method for wearable data; 
1D-DINO, a contrastive learning approach adapted from DINO~\cite{caron2021emerging}; 
SENvT-u4~\cite{okita2023} is trained using multiple pretext tasks, such as Transformer-based masking. Its variant, SENvT-u7~\cite{okita2023}, is trained under a broader multi-task learning framework with additional tasks.
1D-Diffusion Classifier, an adaptation of Diffusion Classifier~\cite{li2023diffusion}; 
and our own DiffFM, a diffusion-based counterpart to FlowFM.

In both transfer learning and fine-tuning settings, \ours{} consistently outperformed all baselines across all five datasets. Under transfer learning, \ours{} improved upon the strong 1D-DINO baseline by up to +7.02\% in accuracy and +6.06\% in F1-score on REALWORLD and WISDM. The strong performance of our diffusion-based DiffFM also validates the effectiveness of our decoupled architecture. In the fine-tuning setting, \ours{} demonstrated substantial gains over the SSL-Wearables baseline, with F1-score improvements ranging from +8.10\% on WISDM to +19.75\% on Opportunity. Figure \ref{fig:pamap_cm} shows confusion matrices for selected results.

To visualize the quality and generalization ability of the learned representations, we used t-SNE~\cite{van2008visualizing} to map signals from the out-of-domain PAMAP2 and REALWORLD datasets into a low-dimensional space (\figref{fig:rep_map}). Despite being trained without labels, the representations form distinct clusters corresponding to activity classes. The effect is particularly clear on REALWORLD, where classes like 'running' and 'standing' form well-separated distributions, and on PAMAP2, where the representations disentangle clusters that overlap in the raw data space.

\begin{figure}[h]
    \begin{center}
    \includegraphics[width=0.5\textwidth]{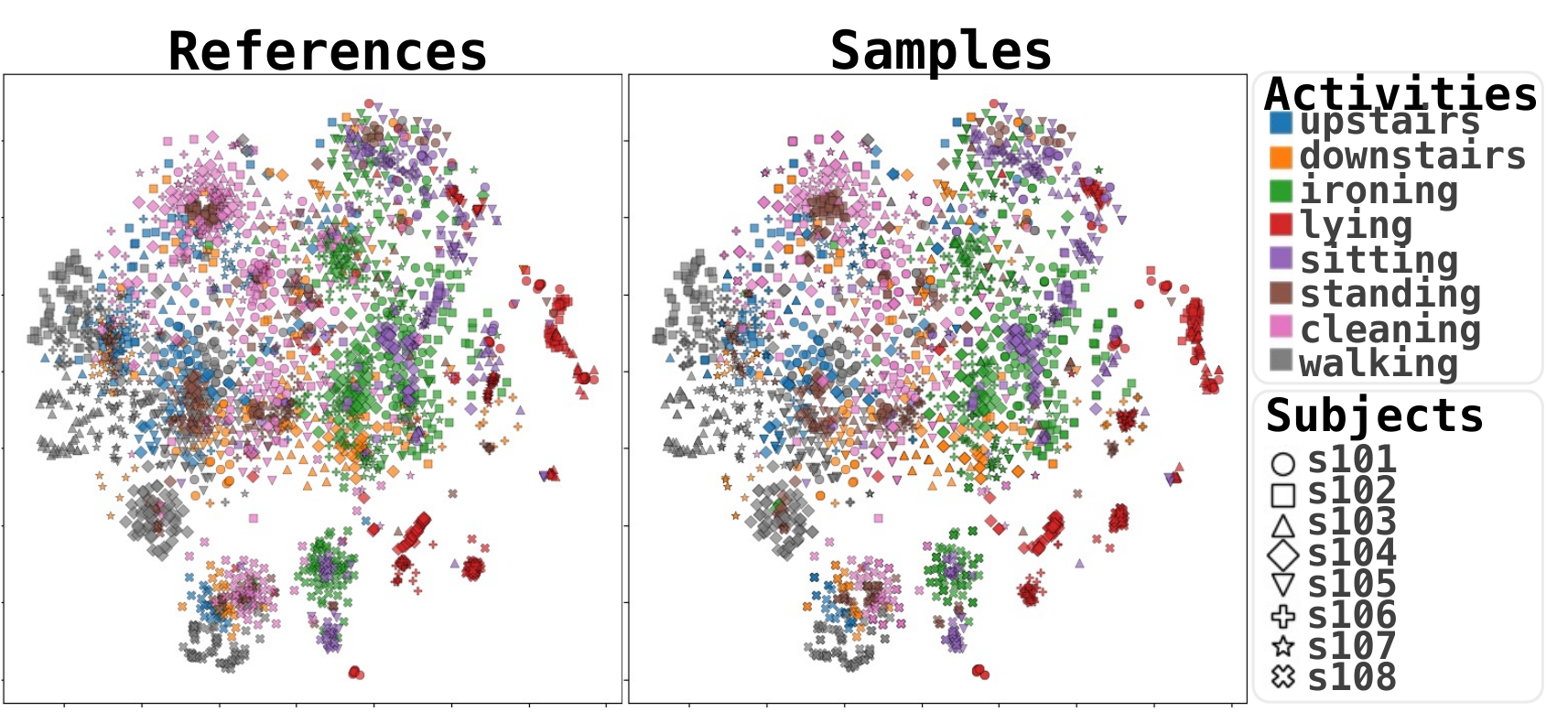}
    \caption{\textbf{t-SNE visualization of generated signals}: Comparison between real data (References) and signals generated from text conditions (Samples) on the PAMAP2 dataset. The generated distribution closely mirrors the real data, demonstrating that the model successfully captures distinct clusters for both activity types (colors) and individual subject characteristics (shapes).}
    \label{fig:ref_sam_map}
    \end{center}
\end{figure}
\subsection{Text-to-Signal Results} \label{section:eval_gen_sensor}
To verify that the learned representations capture rich semantic structures beyond discriminative features, we conducted Text-to-Signal generation task. We constructed text prompts using metadata (e.g., age, sex) and activity labels from the PAMAP2 and REALWORLD datasets. These prompts were converted into condition vectors using the pretrained LLM Llama3 (8B) in a modular text encoder designed for future extensibility.
As shown in \figref{fig:text_to_sensor}, the generated signals clearly reproduce the characteristic patterns of specified actions (e.g., "walking," "lying"). Notably, the model captures subtle differences between similar activities, such as "walking" and "walking downstairs," and even appears to capture variations among individual subjects. This indicates an ability to translate semantic nuances from text into the statistical properties of time-series data.
Furthermore, a t-SNE map of the PAMAP2 dataset (\figref{fig:ref_sam_map}) shows that the distribution of text-generated signals closely mirrors that of the original data, successfully distinguishing between both activities and subjects.

These results qualitatively demonstrate our framework's ability to interpret complex semantic conditions in text to generate high-quality corresponding data. The success of this Text-to-Signal generation opens promising avenues for practical applications, such as synthetic data generation for simulations or conditional data augmentation.

\subsection{Computational Cost Reduction} \label{section:efficient}
\ours{} led to a direct reduction in total energy consumption during the training phase. Through our experiments, we confirmed that flow matching training is more stable and converges faster than diffusion models. This efficiency stems from its approach of directly learning a simpler velocity field via an ODE, bypassing the complex, iterative noising and denoising steps required by diffusion models. In a comparative experiment on training time with an equivalent setup (RTX 4090, batch size 1024), \ours{} (3.2h) succeeded in reducing the training time by 50.4\% compared to DiffFM (6.5h). This indicates that the total energy consumed during the entire training process was halved, which is a significant contribution to the sustainability of large-scale foundation model development.

\begin{table}[h]
\begin{center}
    \caption{\textbf{Generation speed and quality during inference}: FlowFM achieves a 51.0x speedup compared to the diffusion-based DiffFM (1000 steps) while maintaining high generative quality (FID: 5.014), demonstrating significant efficiency gains.}
    \label{tab:inference}
    \scalebox{0.92}{
    \begin{tabular}{c|cccc} \hline\hline
        Model & steps & time[s] / 1 sample$\downarrow$ & FID$\downarrow$ & IS$\uparrow$ \\ \hline
        DiffFM & 1000 & 4.62 $\times 10^{-2}$ & 5.047 & 2.751 \\
        DiffFM &  100 & 4.58 $\times 10^{-3}$ & 8.837 & 2.800 \\
        DiffFM &   50 & 2.29 $\times 10^{-3}$ & 10.08 & \l{2.819} \\ \hline
        FlowFM &   -  & \l{9.05 $\times 10^{-4}$} & \l{5.014} & 2.626 \\ \hline
    \end{tabular}
    }
\end{center}
\end{table}

In the inference phase, \ours{} showed a distinct advantage in generation speed. As shown in \tabref{tab:inference}, \ours{} achieved up to a 51.0x speedup over DiffFM while maintaining high generation quality (FID: 5.014). This speedup is due to the different generative processes: FlowFM solves an ODE, which a numerical solver can compute in far fewer steps than the numerous iterative denoising steps required by diffusion models. Furthermore, compared to generation with the faster DDIM (50-100 steps), it achieved a speed improvement of 2.53-5.06x. This means that the energy consumption per generated sample was dramatically reduced without sacrificing generation quality. This speed improvement highlights its potential contribution to real-time processing and edge AI applications.

\section{Conclusion} \label{section:conclusion}
In this work, we proposed \ours{}, a novel foundation model that applies the generative capabilities of flow matching to representation learning. \ours{} aims to capture the essential structure of data while resolving the computational inefficiency of conventional diffusion models by jointly training a representation encoder, which extracts representations from input, and a flow matching model that generates data conditioned on those representations.

To validate the effectiveness of our proposed method, we conducted evaluations on multiple public datasets using wearable sensor data. The results show that in HAR tasks, \ours{} outperformed all existing SSL methods, including the state-of-the-art SSL-Wearables, across all five datasets. This result quantitatively demonstrates that \ours{}, based on its new principles, can acquire versatile representations applicable to a diverse range of downstream tasks.
Furthermore, in terms of computational efficiency, \ours{} significantly reduced training time compared to the diffusion-based DiffFM, halving the total energy consumption. In the Text-to-Signal generation task, it achieved up to a 51.0x speedup over DiffFM while maintaining high generative quality, demonstrating that energy consumption can be dramatically reduced without sacrificing generative quality.

Based on these results, we conclude that \ours{} is a promising framework that not only overcomes the long-standing trade-off between generative quality and recognition performance but also enhances computational efficiency and quality. These characteristics position \ours{} as a vital step towards realizing on-device foundation models that operate within strict latency and energy budgets. Ultimately, our approach provides a scalable blueprint for sustainable AI, ensuring that the benefits of large-scale representation learning are accessible even in resource-limited edge environments. Future work includes applying this framework to diverse modalities such as images and audio and adapting it to more complex downstream tasks.

\bibliographystyle{plain}
\bibliography{references}

\begin{thebibliography}{10}

\bibitem{bordes2022high}
Florian Bordes, Randall Balestriero, and Pascal Vincent.
\newblock High fidelity visualization of what your self-supervised
  representation knows about.
\newblock {\em arXiv preprint arXiv:2112.09164}, 2021.

\bibitem{bruno2013analysis}
Barbara Bruno, Fulvio Mastrogiovanni, Antonio Sgorbissa, Tullio Vernazza, and
  Renato Zaccaria.
\newblock Analysis of human behavior recognition algorithms based on
  acceleration data.
\newblock pages 1602--1607, 2013.

\bibitem{caron2021emerging}
Mathilde Caron, Hugo Touvron, Ishan Misra, Herv{\'e} J{\'e}gou, Julien Mairal,
  Piotr Bojanowski, and Armand Joulin.
\newblock Emerging properties in self-supervised vision transformers.
\newblock In {\em Proceedings of the IEEE/CVF international conference on
  computer vision}, pages 9650--9660, 2021.

\bibitem{chen2020simple}
Ting Chen, Simon Kornblith, Mohammad Norouzi, and Geoffrey Hinton.
\newblock A simple framework for contrastive learning of visual
  representations.
\newblock In {\em International conference on machine learning}, pages
  1597--1607. PMLR, 2020.

\bibitem{chen2024deconstructing}
Xinlei Chen, Zhuang Liu, Saining Xie, and Kaiming He.
\newblock Deconstructing denoising diffusion models for self-supervised
  learning.
\newblock {\em arXiv preprint arXiv:2401.14404}, 2024.

\bibitem{devlin2019bert}
Jacob Devlin, Ming-Wei Chang, Kenton Lee, and Kristina Toutanova.
\newblock Bert: Pre-training of deep bidirectional transformers for language
  understanding.
\newblock In {\em Proceedings of the 2019 conference of the North American
  chapter of the association for computational linguistics: human language
  technologies, volume 1 (long and short papers)}, pages 4171--4186, 2019.

\bibitem{dombrowski2024trade}
Mischa Dombrowski, Hadrien Reynaud, Johanna~P M{\"u}ller, Matthew Baugh, and
  Bernhard Kainz.
\newblock Trade-offs in fine-tuned diffusion models between accuracy and
  interpretability.
\newblock In {\em Proceedings of the AAAI Conference on Artificial
  Intelligence}, volume~38, pages 21037--21045, 2024.

\bibitem{dosovitskiy2020image}
Alexey Dosovitskiy.
\newblock An image is worth 16x16 words: Transformers for image recognition at
  scale.
\newblock {\em arXiv preprint arXiv:2010.11929}, 2020.

\bibitem{he2022masked}
Kaiming He, Xinlei Chen, Saining Xie, Yanghao Li, Piotr Doll{\'a}r, and Ross
  Girshick.
\newblock Masked autoencoders are scalable vision learners.
\newblock In {\em Proceedings of the IEEE/CVF conference on computer vision and
  pattern recognition}, pages 16000--16009, 2022.

\bibitem{he2019moco}
Kaiming He, Haoqi Fan, Yuxin Wu, Saining Xie, and Ross Girshick.
\newblock Momentum contrast for unsupervised visual representation learning.
\newblock {\em arXiv preprint arXiv:1911.05722}, 2019.

\bibitem{ho2020denoising}
Jonathan Ho, Ajay Jain, and Pieter Abbeel.
\newblock Denoising diffusion probabilistic models.
\newblock {\em Advances in neural information processing systems},
  33:6840--6851, 2020.

\bibitem{huang2024dual}
Lei Huang, Tingyang Xu, Yang Yu, Peilin Zhao, Xingjian Chen, Jing Han, Zhi Xie,
  Hailong Li, Wenge Zhong, Ka-Chun Wong, et~al.
\newblock A dual diffusion model enables 3d molecule generation and lead
  optimization based on target pockets.
\newblock {\em Nature Communications}, 15(1):2657, 2024.

\bibitem{hudson2024soda}
Drew~A Hudson, Daniel Zoran, Mateusz Malinowski, Andrew~K Lampinen, Andrew
  Jaegle, James~L McClelland, Loic Matthey, Felix Hill, and Alexander Lerchner.
\newblock Soda: Bottleneck diffusion models for representation learning.
\newblock In {\em Proceedings of the IEEE/CVF Conference on Computer Vision and
  Pattern Recognition}, pages 23115--23127, 2024.

\bibitem{li2023diffusion}
Alexander~C Li, Mihir Prabhudesai, Shivam Duggal, Ellis Brown, and Deepak
  Pathak.
\newblock Your diffusion model is secretly a zero-shot classifier.
\newblock {\em arXiv preprint arXiv:2303.16203}, 2023.

\bibitem{lipman2022flow}
Yaron Lipman, Ricky~TQ Chen, Heli Ben-Hamu, Maximilian Nickel, and Matt Le.
\newblock Flow matching for generative modeling.
\newblock {\em arXiv preprint arXiv:2210.02747}, 2022.

\bibitem{lipman2024flow}
Yaron Lipman, Marton Havasi, Peter Holderrieth, Neta Shaul, Matt Le, Brian
  Karrer, Ricky~TQ Chen, David Lopez-Paz, Heli Ben-Hamu, and Itai Gat.
\newblock Flow matching guide and code.
\newblock {\em arXiv preprint arXiv:2412.06264}, 2024.

\bibitem{liu2022flow}
Xingchao Liu, Chengyue Gong, and Qiang Liu.
\newblock Flow straight and fast: Learning to generate and transfer data with
  rectified flow.
\newblock {\em arXiv preprint arXiv:2209.03003}, 2022.

\bibitem{meijer2024rise}
Caspar Meijer and Lydia~Y Chen.
\newblock The rise of diffusion models in time-series forecasting.
\newblock {\em arXiv preprint arXiv:2401.03006}, 2024.

\bibitem{okita2023}
Tsuyoshi Okita, Kosuke Ukita, Koki Matsuishi, Masaharu Kagiyama, Kodai Hirata,
  and Asahi Miyazaki.
\newblock Towards llms for sensor data: Multi-task self-supervised learning.
\newblock {\em In Adjunct Proceedings of the 2023 ACM International Joint Conf
  erence on Pervasive and Ubiquitous Computing \& the 2023 ACM International
  Symposium on Wearable Computing}, pages 499--504, 2023.

\bibitem{peebles2023scalable}
William Peebles and Saining Xie.
\newblock Scalable diffusion models with transformers.
\newblock pages 4195--4205, 2023.

\bibitem{reiss2012introducing}
Attila Reiss and Didier Stricker.
\newblock Introducing a new benchmarked dataset for activity monitoring.
\newblock {\em 2012 16th international symposium on wearable computers}, pages
  108--109, 2012.

\bibitem{roggen2010collecting}
Daniel Roggen, Alberto Calatroni, Mirco Rossi, Thomas Holleczek, Kilian
  F{\"o}rster, Gerhard Tr{\"o}ster, Paul Lukowicz, David Bannach, Gerald Pirkl,
  Alois Ferscha, et~al.
\newblock Collecting complex activity datasets in highly rich networked sensor
  environments.
\newblock {\em 2010 Seventh international conference on networked sensing
  systems (INSS)}, pages 233--240, 2010.

\bibitem{rombach2022high}
Robin Rombach, Andreas Blattmann, Dominik Lorenz, Patrick Esser, and Bj{\"o}rn
  Ommer.
\newblock High-resolution image synthesis with latent diffusion models.
\newblock {\em Proceedings of the IEEE/CVF conference on computer vision and
  pattern recognition}, pages 10684--10695, 2022.

\bibitem{shaoul2024multi}
Yorai Shaoul, Itamar Mishani, Shivam Vats, Jiaoyang Li, and Maxim Likhachev.
\newblock Multi-robot motion planning with diffusion models.
\newblock {\em arXiv preprint arXiv:2410.03072}, 2024.

\bibitem{shen2023naturalspeech}
Kai Shen, Zeqian Ju, Xu~Tan, Yanqing Liu, Yichong Leng, Lei He, Tao Qin, Sheng
  Zhao, and Jiang Bian.
\newblock Naturalspeech 2: Latent diffusion models are natural and zero-shot
  speech and singing synthesizers.
\newblock {\em arXiv preprint arXiv:2304.09116}, 2023.

\bibitem{song2020score}
Yang Song, Jascha Sohl-Dickstein, Diederik~P Kingma, Abhishek Kumar, Stefano
  Ermon, and Ben Poole.
\newblock Score-based generative modeling through stochastic differential
  equations.
\newblock {\em arXiv preprint arXiv:2011.13456}, 2020.

\bibitem{sztyler2016body}
Timo Sztyler and Heiner Stuckenschmidt.
\newblock On-body localization of wearable devices: An investigation of
  position-aware activity recognition.
\newblock pages 1--9, 2016.

\bibitem{tong2023improving}
Alexander Tong, Kilian Fatras, Nikolay Malkin, Guillaume Huguet, Yanlei Zhang,
  Jarrid Rector-Brooks, Guy Wolf, and Yoshua Bengio.
\newblock Improving and generalizing flow-based generative models with
  minibatch optimal transport.
\newblock {\em arXiv preprint arXiv:2302.00482}, 2023.

\bibitem{tong2023simulation}
Alexander Tong, Nikolay Malkin, Kilian Fatras, Lazar Atanackovic, Yanlei Zhang,
  Guillaume Huguet, Guy Wolf, and Yoshua Bengio.
\newblock Simulation-free schr$\backslash$" odinger bridges via score and flow
  matching.
\newblock {\em arXiv preprint arXiv:2307.03672}, 2023.

\bibitem{van2008visualizing}
Laurens Van~der Maaten and Geoffrey Hinton.
\newblock Visualizing data using t-sne.
\newblock {\em Journal of machine learning research}, 9(11), 2008.

\bibitem{weiss2019smartphone}
Gary~M Weiss, Kenichi Yoneda, and Thaier Hayajneh.
\newblock Smartphone and smartwatch-based biometrics using activities of daily
  living.
\newblock {\em IEEE Access}, 7:133190--133202, 2019.

\bibitem{willetts2018statistical}
Matthew Willetts, Sven Hollowell, Louis Aslett, Chris Holmes, and Aiden
  Doherty.
\newblock Statistical machine learning of sleep and physical activity
  phenotypes from sensor data in 96,220 uk biobank participants.
\newblock {\em Scientific reports}, 8(1):1--10, 2018.

\bibitem{xiang2023denoising}
Weilai Xiang, Hongyu Yang, Di~Huang, and Yunhong Wang.
\newblock Denoising diffusion autoencoders are unified self-supervised
  learners.
\newblock In {\em Proceedings of the IEEE/CVF International Conference on
  Computer Vision}, pages 15802--15812, 2023.

\bibitem{yang2023diffusion}
Xingyi Yang and Xinchao Wang.
\newblock Diffusion model as representation learner.
\newblock In {\em Proceedings of the IEEE/CVF International Conference on
  Computer Vision}, pages 18938--18949, 2023.

\bibitem{yuan2024self}
Hang Yuan, Shing Chan, Andrew~P Creagh, Catherine Tong, Aidan Acquah, David~A
  Clifton, and Aiden Doherty.
\newblock Self-supervised learning for human activity recognition using 700,000
  person-days of wearable data.
\newblock {\em npj Digital Medicine}, 7(1):1--18, 2024.

\bibitem{zhang20244diffusion}
Haiyu Zhang, Xinyuan Chen, Yaohui Wang, Xihui Liu, Yunhong Wang, and Yu~Qiao.
\newblock 4diffusion: Multi-view video diffusion model for 4d generation.
\newblock {\em Advances in Neural Information Processing Systems},
  37:15272--15295, 2024.

\end{thebibliography}
\end{document}